\def\year{2021}\relax
\documentclass[letterpaper]{article} 
\usepackage{aaai21}  
\usepackage{times}  
\usepackage{helvet} 
\usepackage{courier}  
\usepackage[hyphens]{url}  
\usepackage{graphicx} 
\usepackage{graphicx}  
\usepackage{natbib}  
\usepackage{caption} 
\usepackage{bm}
\usepackage[ruled]{algorithm2e}
\usepackage{amsmath}
\usepackage{mathtools}
\usepackage{multirow}
\usepackage{booktabs}
\usepackage{xcolor}

\title{SMIL: Multimodal Learning with Severely Missing Modality}

\author{
    Mengmeng Ma\textsuperscript{\rm 1}, 
    Jian Ren\textsuperscript{\rm 2}, 
    Long Zhao\textsuperscript{\rm 3}, 
    Sergey Tulyakov\textsuperscript{\rm 2}, 
    Cathy Wu\textsuperscript{\rm 1}, 
    Xi Peng\textsuperscript{\rm 1} \\

}
\affiliations{
    \textsuperscript{\rm 1} University of Delaware,
    \textsuperscript{\rm 2} Snap Inc.,
    \textsuperscript{\rm 3} Rutgers University\\
    
    \{mengma, wuc, xipeng\}@udel.edu, 
    \{jren, stulyakov\}@snap.com, 
    lz311@cs.rutgers.edu

}

\begin{document}

\maketitle

\begin{abstract}
A common assumption in multimodal learning is the completeness of training data, i.e., full modalities are available in all training examples. Although there exists research endeavor in developing novel methods to tackle the incompleteness of testing data, e.g., modalities are partially missing in testing examples, few of them can handle incomplete training modalities. The problem becomes even more challenging if considering the case of severely missing, e.g., 90\% training examples may have incomplete modalities. For the first time in the literature, this paper formally studies multimodal learning with missing modality in terms of flexibility (missing modalities in training, testing, or both) and efficiency (most training data have incomplete modality). Technically, we propose a new method named SMIL that leverages Bayesian meta-learning in uniformly achieving both objectives. To validate our idea, we conduct a series of experiments on three popular benchmarks: MM-IMDb, CMU-MOSI, and avMNIST. The results prove the state-of-the-art performance of SMIL over existing methods and generative baselines including autoencoders and generative adversarial networks. Our code is available at https://github.com/mengmenm/SMIL.


\end{abstract}

\section{Introduction}
Multimodal learning attracts intensive research interest because of broad applications such as intelligent tutoring~\cite{petrovica2017emotion}, robotics~\cite{noda2014multimodal}, and healthcare~\cite{frantzidis2010classification}. Generally speaking, existing research efforts mainly focus on how to fuse multimodal data effectively~\cite{liu2018efficient,tensoremnlp17} and how to learn a good representation for each modality~\cite{Tian2020Contrastive}.

\begin{figure}[t]
    \centering
    \includegraphics[width=\linewidth]{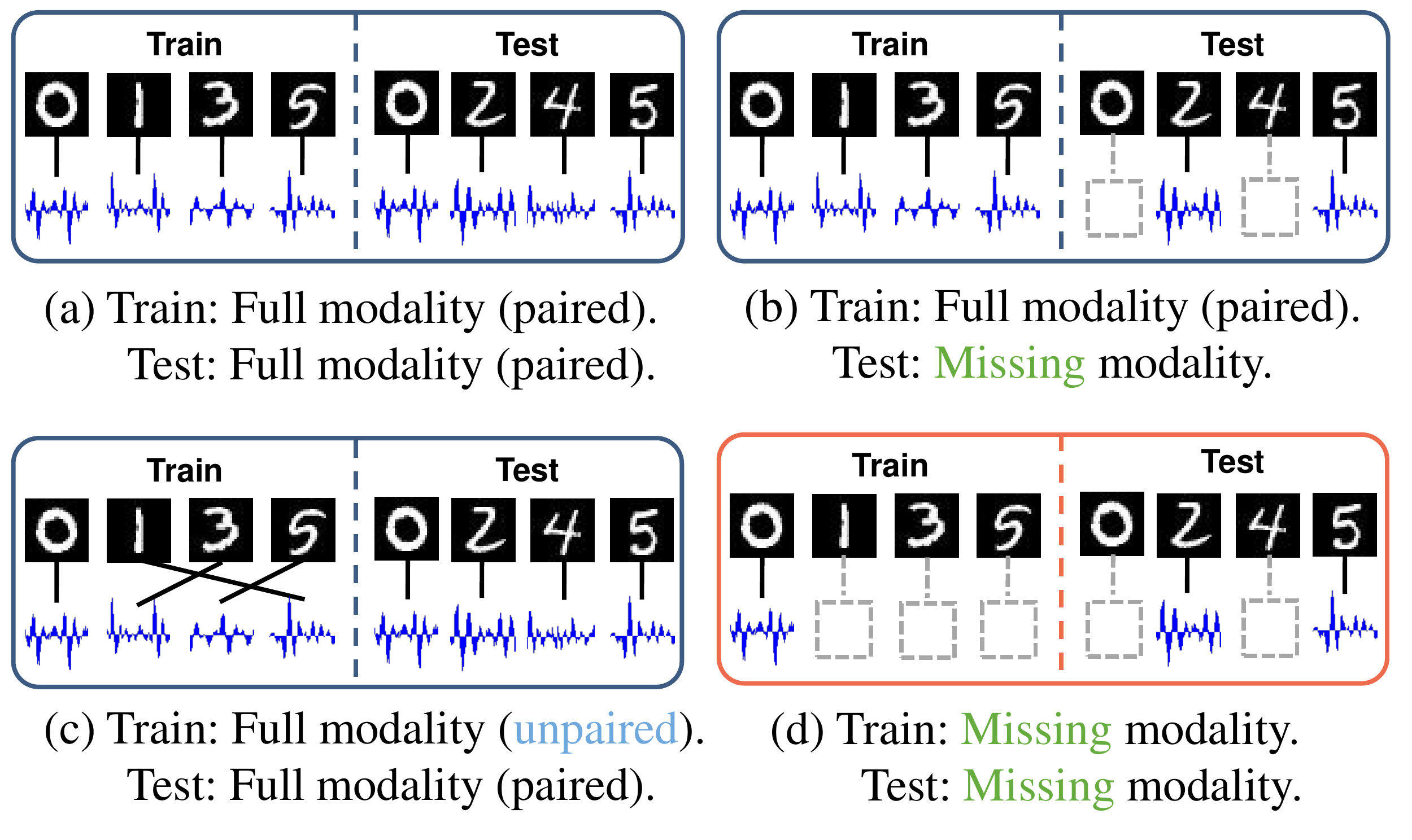}
    \caption{Multimodal learning configurations. {(a)} Train and test with full and paired modality~\cite{ngiam2011multimodal}; {(b)} Testing with missing modality~\cite{tsai2018learning}; {(c)} Training with unpaired modality~\cite{shi2020relating}; {(d)} We study the most challenging configurations of severely missing modality in training, testing, or both.}
    \label{fig:concept}
\end{figure}

A common assumption underlying multimodal learning is the completeness of modality as illustrated in Figure~\ref{fig:concept}. Existing methods~\cite{ngiam2011multimodal,zadeh2017tensor,hou2019deep} often assume full and paired modalities are available in both training and testing data. However, such an assumption may not always hold in real world due to privacy concerns or budget limitations. For example, in social network, we may not be able to access full-modality data since users would apply various privacy and security constraints. In autonomous driving, we may collect many imaginary data but not as so for 3D point cloud because LiDARs are much less affordable than cameras.  

Although there exist a bunch of research efforts~\cite{tsai2018learning,pham2019found} in developing novel methods to tackle the incompleteness of testing data, few of them can handle incomplete training modalities. An interesting yet challenging research question then arises: Can we learn a multimodal model from an incomplete dataset while its performance should as close as possible to the one that learns from a full-modality dataset?

In this paper, we systematically study this problem by proposing multimodal learning with severely missing modality (SMIL). We consider an even more challenging setting that the missing ratio can be as much as $90\%$. More specifically, we design two objectives for SMIL: {\it flexibility} and {\it efficiency}. The former requires our model to uniformly tackle three different missing patterns in training, testing, or both. The latter enforces our model to effectively learn from incomplete modality as fast as possible.

To jointly achieve both objectives, we leverage Bayesian meta-learning framework in designing a new method. The key idea is to perturb the latent feature space so that embeddings of single modality can approximate ones of full modality. We highlight that our method is better than typical generative designs, such as {\it Autoencoder (AE)}~\cite{Tran_2017_CVPR}, {\it Variational Autoencoder (VAE)}~\cite{kingma2013auto}, or {\it Generative Adversarial Network (GAN)}~\cite{goodfellow2014generative}, since they often require a significant amount of full-modality data to learn from, which is usually not available in severely missing modality learning. To summarize, our contribution is three-fold:

\begin{itemize}
\item To the best of our knowledge, we are the first work to systematically study the problem of multimodal learning with severely missing modality.
\item We propose a Bayesian meta-learning based solution to uniformly achieve the goals of {\it flexibility} (missing modalities in training, testing, or both) and {\it efficiency} (most training data have incomplete modality).
\item Extensive experiments on MM-IMDb, CMU-MOSI, and avMNIST validate the state-of-the-art performance of SMIL over generative baselines including AE and GAN. 

\end{itemize}

\section{Related Work}
\begin{figure*}[t]
    \centering
    \includegraphics[width=1.0\linewidth]{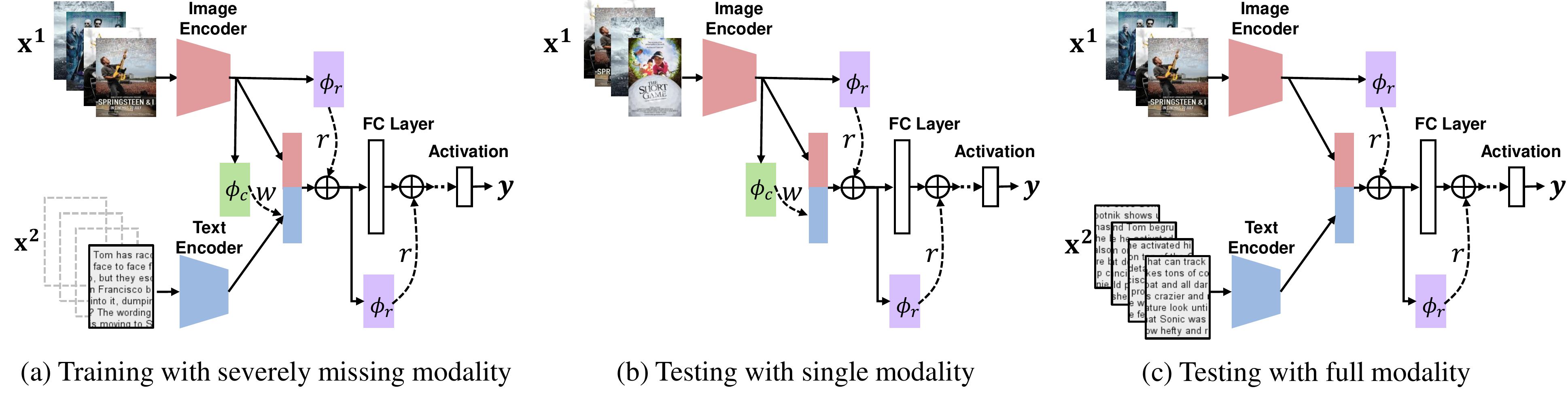}
    \caption{SMIL can uniformly learn from severely missing modality and test with either single or full modality. {\it The reconstruction network $\phi_{c}$} outputs a posterior distribution, from which we sample {\it weight $\omega$} to reconstruct the missing modality using modality priors. {The regularization network $\phi_{r}$} also outputs a posterior distribution, from which we sample {\it regularizer $r$} to perturb latent features for smooth embedding. The collaboration ($\phi_{c}$ and $\phi_{r}$) guarantees flexible and efficient learning.}
    \label{fig:method}
\end{figure*}

{\bf Multimodal learning.}  
Multimodal learning utilizes complementary information contained in multimodal data to improve the performance of various computer vision tasks. One important direction in this area is multimodal fusion, which focuses on effective fusion of multimodal data. Early fusion is a common method which fuses different modalities by feature concatenation, and it has been widely adopted in previous studies~\cite{wang2017select,poria2016convolutional}. Instead of concatenating features, Zadeh~\textit{et al.}~\cite{zadeh2017tensor} proposed a product operation to allow more interactions among different modalities during the fusion process. Liu~\textit{et al.}~\cite{liu2018efficient} utilized modality-specific factors to achieve efficient low-rank fusion. 

Recently, there have been a wide range of research interests in handling missing modalities for multimodal learning, such as testing-time modality missing~\cite{tsai2018learning} and learning with data from unpaired modalities~\cite{shi2020relating}. In this paper, we tackle a more challenging and novel multimodal-learning setting where both training and testing data contain samples that have missing modalities. Generative approaches, such as auto-encoders~\cite{Tran_2017_CVPR,lee2019audio}, GANs~\cite{goodfellow2014generative}, and VAEs~\cite{kingma2013auto}, 
offer a straightforward solution to handle this setting, but these methods are neithor flexible nor efficient as SMIL.

{\bf Meta-regularization.}
Meta-learning algorithms focus on designing models that are able to learn new knowledge and adapt to novel environments quickly with only a few training samples. Previous methods studied meta-learning from the perspective of metric learning~\cite{koch2015siamese,vinyals2016matching,sung2018learning,snell2017prototypical} or probabilistic modeling~\cite{fe2003bayesian,lawrence2004learning}. Recent advances in optimization-based approaches have evoked more interests in meta-learning. MAML~\cite{finn2017model} is a general optimization algorithm designed for few-shot learning and reinforcement learning. It is compatible with models that learn through gradient descent. Nichol~\textit{et al.}~\cite{nichol2018first} further improved the computation efficiency of MAML. Other works adapted MAML for domain generalization~\cite{li2018learning,qiao2020learning} and knowledge distillation~\cite{zhao2020knowledge}. In this work, we extend MAML by learning two auxiliary networks for missing modality reconstruction and feature regularization.

Conventional handcrafted regularization techniques~\cite{hoerl1970ridge,tibshirani1996regression} regularize model parameters to avoid overfitting and increase interpretability.
Balaji~\textit{et al.}~\cite{balaji2018metareg} modeled the regularization function as an additional network learned through meta-learning to regularize model parameters. Li~\textit{et al.}~\cite{li2019feature} followed the same idea of~\cite{balaji2018metareg} but learned an additional network to regularize latent features. Lee~\textit{et al.}~\cite{Lee2020Meta} proposed a more general algorithm for latent feature regularization. Other than perturbing features, we propose to learn the regularization function following~\cite{Lee2020Meta} but regularize the feature to reduce discrepancy between the reconstructed and true modality. 


{\bf Multimodal generative models.}  Generative models for multimodal learning fall into two categories: cross-modal generation and joint-model generation. Cross-modal generation methods, such as conditional VAE (CVAE)~\cite{sohn2015learning} and conditional multimodal auto-encoder~\cite{,pandey2017variational}, learn a conditional generative model over all modalities. On the other hand, joint-model generation approaches learn the joint distribution of multimodal data. Multimodal variational autoencoder (MVAE)~\cite{wu2018multimodal} models the joint posterior as a product-of-expert (PoE). Multimodal VAE (JMVAE)~\cite{suzuki2016joint} learns a shared representation with a joint encoder. With only a few modifications to the original algorithms, we show that multimodal generative models serve as strong baselines for learning with severely missing modalities proposed in this paper.

\section{Proposed Method}
\label{sec:method}
We are interested in multimodal learning with severely missing modality, e.g., 90\% of the training samples contain incomplete modalities. In this paper, without loss of generality, we consider a multimodal dataset containing two modalities. Formally, we let $\mathcal{D}=\{\mathcal{D}^f,\mathcal{D}^m\}$ denote a multimodal dataset; $\mathcal{D}^f=\{\boldsymbol{\mathbf{x}}_i^{1},\mathbf{x}_i^{2},y_i \}_{i}$ is a modality-complete dataset, where $\mathbf{x}_i^{1}$ and $\mathbf{x}_{i}^{2}$ represent two different modalities of $i$-th sample and $y_i$ is the corresponding class label; $\mathcal{D}^m=\{ \boldsymbol{\mathbf{x}}_j^{1}, y_j \}_{j}$ is a modality-incomplete dataset, where one modality is missing. Our target is to leverage both modality-complete and modality-incomplete data for model training. We propose to address this problem from two perspectives: {\it 1) Flexibility:} how to uniformly handle missing modality in training, testing, or both? {\it 2) Efficiency:} how to improve training efficiency when major data suffers from missing modality?


{\bf Flexibility.} 
We aim to achieve a unified model that can handle missing modality in training, testing, or both.
Our idea is to employ a feature reconstruction network to achieve this goal. Instead of following the conventional data reconstruction approaches~\cite{lee2019audio,Tran_2017_CVPR}, the feature reconstruction network will leverage the available modality to generate an approximation of the missing-modality feature in a highly efficient way. This will generate complete data in the latent feature space and facilitate the flexibility in two aspects. On the one hand, our model can excavate the full potential of hybrid data by using both modality-complete and -incomplete data for joint training. On the other hand, when testing, by turning on or off the feature reconstruction network, our model can tackle modality-incomplete or -complete inputs in a unified manner.


{\bf Efficiency.}
We intend to train a model on the modality severely missing dataset to achieve comparable performance as the model trained on a full-modality dataset. However, the severely missing modality setting poses significant learning challenges to the feature reconstruction network. The network would be highly bias-prone due to the scarcity of modality-complete data, yielding degraded and low-quality feature generations. Directly train a model with degraded and low-quality features will hinder the efficiency of the training process. We propose a feature regularization approach to address this issue. The idea is to leverage a Bayesian neural network to assess the data uncertainty by performing feature perturbations. The uncertainty assessment is used as feature regularization to overcome model and data bias. Compared with previous deterministic regularization approaches~~\cite{balaji2018metareg,zhao2020knowledge}, the proposed uncertainty-guided feature regularization will significantly improve the capacity of the multimodal model for robust generalization behaviors in tackling severely incomplete data.


{\bf A meta-learning framework.} 
To effectively organize model training, we integrate the main network $f_{\theta}$ parameterized by $\bm \theta$, the reconstruction network $f_{\phi_{c}}$ parameterized by $\bm \phi_c$, and the regularization network $f_{\phi_{r}}$ parameterized by $\bm \phi_{r}$ in a modified \textit{Model-Agnostic Meta-Learning (MAML)}~\cite{finn2017model} framework. An overview of our learning framework is shown in Figure~2. In the following sections, we describe the implementation of the feature reconstruction and regularization network.

\subsection{Missing Modality Reconstruction}
\label{subsec:missingModality}
We introduce the feature reconstruction network to approximate the missing modality. For a modality-incomplete sample, the missing modality is reconstructed conditioned on the available modality. Given the observed modality $\mathbf{x}^{1}$, in order to obtain the reconstruction $\hat{\mathbf{x}}^{2}$ of the missing modality, we optimize the following objective for the reconstruction network:
\begin{equation}
\label{eqn:generation}
\bm \phi_c^{*}{=}\,\underset{\bm \phi_c}{\arg \min}\, \mathbf E_{p(\hat{\mathbf{x}}^{1},\,\mathbf{x}^{2})}(-\mathrm{log}\, p(\hat{\mathbf{x}}^{2}| \mathbf{x}^{1};\bm \phi_c)).
\end{equation}
However, under severely missing modality, it is non-trivial to train a reconstruction network from limited modality-complete samples. 
Inspired by~\cite{kuo2019shapemask}, we approximate the missing modality using a weighted sum of modality priors learned from the modality-complete dataset. In this case, the reconstruction network are trained to predict weights of the priors instead of directly generating the missing modality. We achieve this by learning a set of modality priors $\mathcal{M}$ which can be clustered among all modality-complete samples using K-means~\cite{macqueen1967kmeans} or PCA~\cite{pearson1901pca}.

Specifically, let $\bm \omega$ represent the weights assigned to each modality prior. We model $\bm \omega$ as a multivariate Gaussian with fixed means and changeable variances as $\mathcal{N}(\textbf{I}, \bm \sigma)$. The variances are predicted by the feature reconstruction network $\bm \sigma$ = $f_{\phi_{c}}(\bm  {\mathbf{x}^1})$. 
Given the weights $\bm \omega$, we can reconstruct the missing modality $\hat{\mathbf{x}}^2$ by calculating the weighted sum of the modality priors. Then, the reconstructed missing modality can be achieved by:
\begin{equation} 
    \hat{\mathbf{x}}^2= \left \langle \bm \omega , \mathcal{M} \right\rangle, \mathrm{where}\; \bm \omega \sim \mathcal{N}(\textbf{I}, \bm \sigma).
\end{equation}
We note that modeling $\bm \omega$ as multivariate random variables introduces randomness and uncertainty to the reconstruction process, which has been proved to be beneficial in learning sophisticated distributions~\cite{Lee2020Meta}.

\subsection{Uncertainty-Guided Feature Regularization}
\label{subsec:regularization}
We propose to regularize the latent features by a feature regularization network. In each layer, the regularization network takes the features of the previous layer as input and applies regularization to the features of the current layer. Let $\mathbf{r}$ denote the generated regularization and $\mathbf{h}^l$ be the latent feature of the $l$-th layer. Instead of generating a deterministic regularization $\mathbf{r}=f_{\phi_{r}}(\mathbf{h}^{l-1})$, we assume that $\mathbf{r}$ follows a multivariate Gaussian distribution $\mathcal{N}(\bm \mu, \bm \sigma)$, where the means and variances are calculated using $(\bm \mu, \bm \sigma)$ = $f_{\phi_{r}}(\mathbf{h}^{l-1})$. Then, we can compute the regularized feature by the following equation:
\begin{equation}
 \mathbf{h}^{l} \coloneqq \mathbf{h}^l \circ \mathrm{Softplus}(\mathbf{r}),\; \mathrm{where}\;  \mathbf{r} \sim \mathcal{N}( \bm \mu, \bm \sigma),\label{eqn:reg}
\end{equation}
where $\circ$ is a predefined operation (either addition or multiplication) for feature regularization. In our experiments, we observe that directly applying regularization to latent features will prevent the feature regularization network from convergence. Hence, we adopt $\mathrm{Softplus}$~\cite{dugas2000incorporating} activation to weaken the regularization.

\subsection{A Bayesian Meta-Learning Framework} 
\label{subsec:bayesianInference}
We leverage a Bayesian Meta-Learning framework to jointly optimizing all the networks. Specifically, we \textit{meta-train} the main network $f_\theta$ on $D^m$ with the help of reconstruction $f_{\phi_c}$ network and regularization $f_{\phi_r}$ network. Then, we \textit{meta-test} the updated main network $f_{\theta^*}$ on $D^f$. Finally, we \textit{meta-update} network parameters $\{ \bm \theta, \bm \phi_c, \bm \phi_r \}$ by gradient descent.

For simplicity, we let $\bm \psi$ = \{$\bm \phi_{c},\bm \phi_{r}\}$ denote the combination of the parameters of the reconstruction and regularization network. Our framework aims to optimize the following objective function:
\begin{equation}
\begin{gathered}
\label{eqn:objective}
\min\limits_{\bm \theta,\bm \psi } \mathcal{L}(\mathcal{D}^{f};\bm\theta^{*}, \bm \psi), \\
\mathrm{where}\; \bm \theta^{*}{=}\,\bm \theta - \alpha\nabla_{\bm \theta} \mathcal{L}(\mathcal{D}^{m};\bm \psi).
\end{gathered}
\end{equation}
For the above function, $ \mathcal{L}$ denotes the empirical loss such as cross entropy, and $\alpha$ is the inner-loop step size.

We use $\textbf{X}$ and $\textbf{Y}$ to represent all training samples and their corresponding labels, respectively. Let $\mathbf{z}= \{ \bm \omega, \mathbf{r}\}$ be the collection of the generated weights and regularization.
Then, inspired by~\cite{finn2018probabilistic,gordon2018metalearning,Lee2020Learning}, we define the generative process as optimizing the likelihood in a meta-learning framework:
\begin{equation}
\begin{aligned}
    p(\textbf{Y}, \textbf{z}|\textbf{X};\bm \theta)
    =p(\textbf{z})
    \prod_{i=1}^{N}p({\boldsymbol{y}}_{i}|{\boldsymbol{\mathbf{x}}}_{i}^1,\boldsymbol{\mathbf{x}}_{i}^2, \textbf{z};\bm \theta)
    \prod_{j=1}^{M}p({\boldsymbol{y}}_{j}|{\boldsymbol{\mathbf{x}}}_{j}^1,\textbf{z};\bm \theta).\label{eqn:p}        
\end{aligned}
\end{equation}

The goal of Bayesian Meta-Learning is to maximize the conditional likelihood: $\mathrm{log}\, p(\mathbf{Y}|\mathbf{X};\bm \theta)$. 
However, solving it involves the true posterior $p(\textbf{z}|\textbf{X})$, which is intractable. Instead, we approximate the true posterior distribution by an amortized distribution $q(\textbf{z}|\textbf{X};\bm \psi)$~\cite{finn2018probabilistic,gordon2018metalearning,Lee2020Learning}.
The resulting form of approximated lower bound for our meta-learning framework can be defined as:
\begin{multline}
\label{eq:elbo}
    \mathcal{L}_{\bm \theta,\bm \psi}
    =\bm {E}_{q(\textbf{z}|\textbf{X};\bm \theta, \bm \psi)}[\mathrm{log}\, p(\mathbf{Y}|\mathbf{X},\textbf{z};\bm \theta)]-\\
    \mathrm{KL}[q(\textbf{z}|\textbf{X};\bm \psi) \| p(\textbf{z}|\textbf{X})]. 
\end{multline}
We maximize this lower bound by Monte-Carlo (MC) sampling. After combining all these together, we obtain the full training objective of the proposed meta-learning framework for $\bm \theta$ and $\bm \psi$ which is defined as:
\begin{equation} \label{eqn:6}
\begin{gathered}
\min_{\bm \theta,\bm \psi} \frac{1}{L}\sum^{L}_{l=1} - \mathrm{log}\, p(\boldsymbol{y}_{j}|{\mathbf{x}}_{j}^1,{\mathbf{x}}_{j}^2, \text{z}_{l};\bm \theta) + \mathrm{KL}[q(\textbf{z}|\textbf{X};\bm \psi) \| p(\textbf{z}|\textbf{X})]\\
\mathrm{with}\; \mathbf{z}_{l} \sim q(\mathbf{z}|\mathbf{X};\bm \psi), 
\end{gathered}
\end{equation} 
where $L$ is the number of MC sampling. We show our detailed algorithm in Algorithm~\ref{alg:alg}.

\begin{algorithm}[t]
	\caption{Bayesian Meta-Learning Framework.}\label{alg:alg}
	\LinesNumbered
	\label{alg:overrall}
	\KwIn{Multimodal dataset $\mathcal{D}=\{D^{f},D^{m}\}$; \# of iterations K; inner learning rate $\alpha$; outer learning rate $\beta$.}
	\BlankLine
	\While{not converged}{
	    Sample $ \{\mathbf{x}_j^1, y_j \} \sim D^{m}; \{\mathbf{x}_i^1, \mathbf{x}_i^2, y_i \} \sim D^{f}$ \\
	    $\bm \theta_{0} \leftarrow \bm \theta$\\
	    \textbf{Meta-train}: \\
	    \For{$k=0\; \mathrm{to}\; K-1$}{
	    Sample $\boldsymbol{\tilde{\mathbf{z}}}_j \sim p(\boldsymbol{\mathbf{z}}_{j}|\boldsymbol{\mathbf{x}}^{1}_{j};\bm \psi, \bm \theta_k)$ \\
	    $\bm \theta_{k+1} \leftarrow \bm \theta_{k} - \alpha \nabla_{\bm \theta_{k}} [-\mathrm{log}\, p(\boldsymbol{y}_{j}|\boldsymbol{\mathbf{x}}^{1}_{j},\boldsymbol{\tilde{\mathbf{z}}}_{j};\bm \theta_{k})]$ 
	    }
        $\bm \theta^{*} \leftarrow \bm \theta_{K}$  \\
	    \textbf{Meta-test \& Meta-update:} \\
	    $\bm \theta \leftarrow \bm \theta - \beta \nabla_{\bm \theta} [-\mathrm{log}\, p(\boldsymbol{y}_{i}|\boldsymbol{\mathbf{x}}^{1}_{i}, \boldsymbol{\mathbf{x}}^{2}_{i},\boldsymbol{\tilde{z}}_{i};\bm \theta^{*})]$  \\
	    $\bm \psi \leftarrow \bm \psi - \beta \nabla_{\bm \psi} [-\mathrm{log}\, p(\boldsymbol{y}_{i}|\boldsymbol{\mathbf{x}}^{1}_{i}, \boldsymbol{\mathbf{x}}^{2}_{i},\boldsymbol{\tilde{z}}_{i};\bm \theta^{*})]$ \\
	   
	}
\end{algorithm}

	   

\section{Experiments}
In this section, we analyze the results of the proposed algorithm for multimodal learning with severely missing modality on three datasets from two perspectives: \textit{efficiency under severely missing modality} (Section~4.2) and \textit{flexibility to various modality missing pattern} (Section~4.3).

\subsection{Experiment Setting}
{\bf Datasets.} Totally three datasets are used in the experiment:
\begin{itemize}


\item \textit{The Multimodal IMDb (MM-IMDb)}~\cite{arevalo2017gated} contains two modalities: image and text. We conduct experiments on this dataset to predict a movie genre using image or text modality, which is a multi-label classification task as multiple genres could be assigned to a single movie. The dataset includes $25,956$ movies and $23$ classes. We follow the training and validation splits provided in the previous work~\cite{vielzeuf2018centralnet}.

\item \textit{CMU Multimodal Opinion Sentiment Intensity~(CMU-MOSI)}~\cite{zadeh2016mosi} consists of $2,199$ opinion video clips from YouTube movie reviews. Each clip contains three modalities: the image modality includes the visual gesture, the text modality includes the transcribed speech, and the audio modality includes the automatic audio. We use the feature extraction model from~\citet{liu2018efficient} for each modality. We conduct experiments on this dataset to predict the sentiment class of the clips, which is a binary classification task as the sentiment of video clips can be either negative or positive. There are $1,284$ segments in the training set, $229$ in the validation set, and $686$ in the test set. In the experiment section, we only use the image and text modality. 

\item  \textit{Audiovision-MNIST (avMNIST)}~\cite{vielzeuf2018centralnet} consists of an independent image and audio modalities. The images, which are digits from $0$ to $9$, are collected from the MNIST dataset~\cite{lecun1998gradient} with a size of $28\times28$, and the audio modality is collected from Free Spoken Digits Dataset \footnote{\url{https://github.com/Jakobovski/free-spoken-digit-dataset}} containing raw $1,500$ audios. We use the {mel-frequency cepstral coefficients} (MFCCs)~\cite{tzanetakis2002musical} as the representation of audio modality. Each raw audio is processed by MFCCs to get a sample with a size of $20\times20\times1$. 
The dataset contains $1,500$ samples for both image and audio modalities. We randomly select $70$\% data for training and use the rest for validation.
\end{itemize}


\begin{table}[t]
\centering
\resizebox{\linewidth}{!}{
    \begin{tabular}{lcccccc}
    \toprule
    \multirow{2}{*}{Method} & \multicolumn{3}{c}{Accuracy (\%) $\uparrow$} & \multicolumn{3}{c}{F1 Score $\uparrow$}  \\ 
    \cmidrule{2-4}
    \cmidrule{5-7}
          & 10\% & 20\% &100\% &10\% & 20\% & 100\%     \\
    \midrule
    Lower-Bound  & -- & -- &44.8 & -- & -- &27.7\\
    Upper-Bound  & -- & -- &71.0 & -- & -- &70.5 \\
    MVAE & -- &--  &58.5 & -- & -- &58.1\\
    \midrule
    AE & 56.4 &60.4  &--&54.4 & 59.0 &--\\
    GAN &56.5  &60.6  &--&54.6 &59.1 &--\\
    \midrule
    SMIL     & \textbf{60.7} & \textbf{63.3} &--&\textbf{58.0} &\textbf{62.5}&--\\
    \bottomrule
    \end{tabular}}
    \caption{Binary classification accuracy (\%) and F1 Score for different methods under three text modality ratios (10\%, 20\%, and 100\%) on the \textit{CMU-MOSI} dataset. }\label{tab:CMU-MOSI}
\end{table}

{\bf Evaluation metrics.} For MM-IMDb dataset, we follow previous works~\cite{arevalo2017gated,vielzeuf2018centralnet} by adopting the F1 Samples and F1 Micro to evaluate multi-label classification. For CMU-MOSI, we follow~\citet{liu2018efficient} to compute the binary classification accuracy and F1 Score. For avMNIST dataset, we compute accuracy to measure the performance.

{\bf Baseline methods.} We compare the proposed approach with the following baseline methods:
\begin{itemize}
\item \textit{Lower-Bound} is a model trained using single modality of the data, \textit{i.e.,} $100$\% image, $100$\% text, etc. It serves as the lower bound for our method.
\item \textit{Upper-Bound} is a model trained leveraging all modalities of the data, \textit{i.e.,} $100$\% images and $100$\% text, etc. We regard it as the upper bound.
\item  \textit{AE} (Autoencoder)~\cite{lee2019audio} is a deep model used for efficient data encoding. We can use AE to preprocess the original dataset to tackle the severely missing modality problem. We now describe the procedure for preprocessing. First, we sample a dataset containing only modality-complete samples from the original dataset. Then, we assume one modality is missing and train AE to reconstruct the missing modality. Finally, we impute the missing modality of modality-incomplete data using the trained AE. After finishing the imputation, the dataset is now available for multimodal learning. 
\item \textit{GAN} (Generative adversarial network) is a deep generative model composed of a generator and a discriminator. We leverage GAN to tackle our problem following the same procedure as described in AE. 
\item \textit{MVAE}~\cite{wu2018multimodal} is proposed for multimodal generative task. We adopt the widely used linear evaluation protocol to adapt MVAE for classification. Specifically, we first train MVAE using all the modalities. We then keep the learned MVAE frozen to train a randomly initialized linear classifier using the latent representation generated by the encoder of MVAE.
\end{itemize}


\begin{table}[t]
\centering
\resizebox{\linewidth}{!}{
    \begin{tabular}{lcccccc}
    \toprule
    \multirow{2}{*}{Method} & \multicolumn{3}{c}{F1 Samples $\uparrow$} & \multicolumn{3}{c}{F1 Micro $\uparrow$}  \\ 
    \cmidrule{2-4}
    \cmidrule{5-7}
          & 10\% & 20\% &100\% &10\% & 20\% & 100\%     \\ 
    \midrule
    Lower-Bound & -- & -- &47.6& -- & -- &48.2\\
    Upper-Bound  & -- & -- &61.7& -- & -- &62.0 \\
    MVAE & -- &--  &48.4& --& -- &48.6\\
    \midrule
    AE & 44.5 &50.9  &--&44.8& 50.7 &--\\
    GAN & 45.0 &51.1  &--&44.6&51.0 &--\\
    \midrule
    SMIL     & \textbf{49.2} & \textbf{54.1} &--&\textbf{49.5} &\textbf{54.6}&--\\
    \bottomrule
    \end{tabular}}
    \caption{ Multi-label classification scores (F1 Samples and F1 Micro) for different methods under three text modality ratios (10\%, 20\%, and 100\%) on the \textit{MM-IMDb} dataset.}\label{tab:MM-IMDb}
\end{table}

\subsection{Efficiency with Severely Missing Modality}
\textbf{Conclusion:} \textit{Our method demonstrates consistent efficiency, across different datasets, when training data contains a different ratios of modality missing.}

{\bf Setting of missing modality.}
We evaluate the efficiency of our algorithm on two datasets: MM-IMDb and CMU-MOSI. In both datasets, modalities are incomplete for some samples. We define the text modality ratio as $\eta=\frac{M}{N}$, where $ M $ is the number of samples with text modality and $ N $ is the size of overall samples. $\eta$ indicates the severity of modality missing. The smaller of $\eta$, the severer the modality is missing. For both datasets, we assume image modality to be complete, and the text modality to be incomplete. We express all available data points in the form of $100$\% Image + $\eta$\% Text for both datasets.

{\bf Implementation details.} \textbf{CMU-MOSI.} We follow~\citet{liu2018efficient} to get features for the image and text modality. We use three fully-connected (FC) layers with dimension $16$ to get the embedding of image modality. One layer LSTM~\cite{hochreiter1997long} extracts the embedding for text modality. The concatenated feature of two modalities is then fed to FC layers for classification. For training process, we use Adam~\cite{kingma2014adam} optimizer with a batch size of $32$ and train the networks for $5,000$ iterations with a learning rate of $10^{-4}$ for both inner-loop and outer-loop of meta-learning.
\textbf{MM-IMDB.} For image and text modalities, we adopt the feature extraction models from~\citet{arevalo2017gated}. We feed the feature from each modality to a FC layer to align their output dimension. On top of it, we fuse the feature together and send it to FC layers to conduct multi-label classification. We apply Adam optimizer with a batch size of $128$. We train the models for $10,000$ iteration with a learning rate of $10^{-4}$ for inner-loop and $10^{-3}$ fro outer-loop.
Besides, we follow previous work~\cite{vielzeuf2018centralnet} to add a weight of $2.0$ on the positive label to balance the precision and recall since the labels are unbalanced.

{\bf Different ratios of modality missing.} 
The results on CMU-MOSI are shown in Table 1. As can be seen, our approach significantly outperforms all baselines among all ratios of modality missing, which showcases the efficiency of our approach in the missing modality problem. The results also show that the severer the missing modality is, the more efficient our approach is. More specifically, when $\eta$ is $20$\%, our approach outperforms AE and GAN around $5.0$\%, while the improvements increase to $7.6$\% and $7.4$\%, respectively, when $\eta$ decreases to $10$\%. Moreover, our improvements are also consistent on MM-IMDb, as shown in Table~\ref{tab:MM-IMDb}. The improvement increases  as the modality ratio decreasing. From Table~\ref{tab:MM-IMDb}, we see that our approach performs better than all baseline method under different text ratio. Our method outperforms Lower-Bound and MVAE by a large margin, and quite close to Upper-Bound.

\begin{figure}[t]
    \centering
    \includegraphics[width=\linewidth]{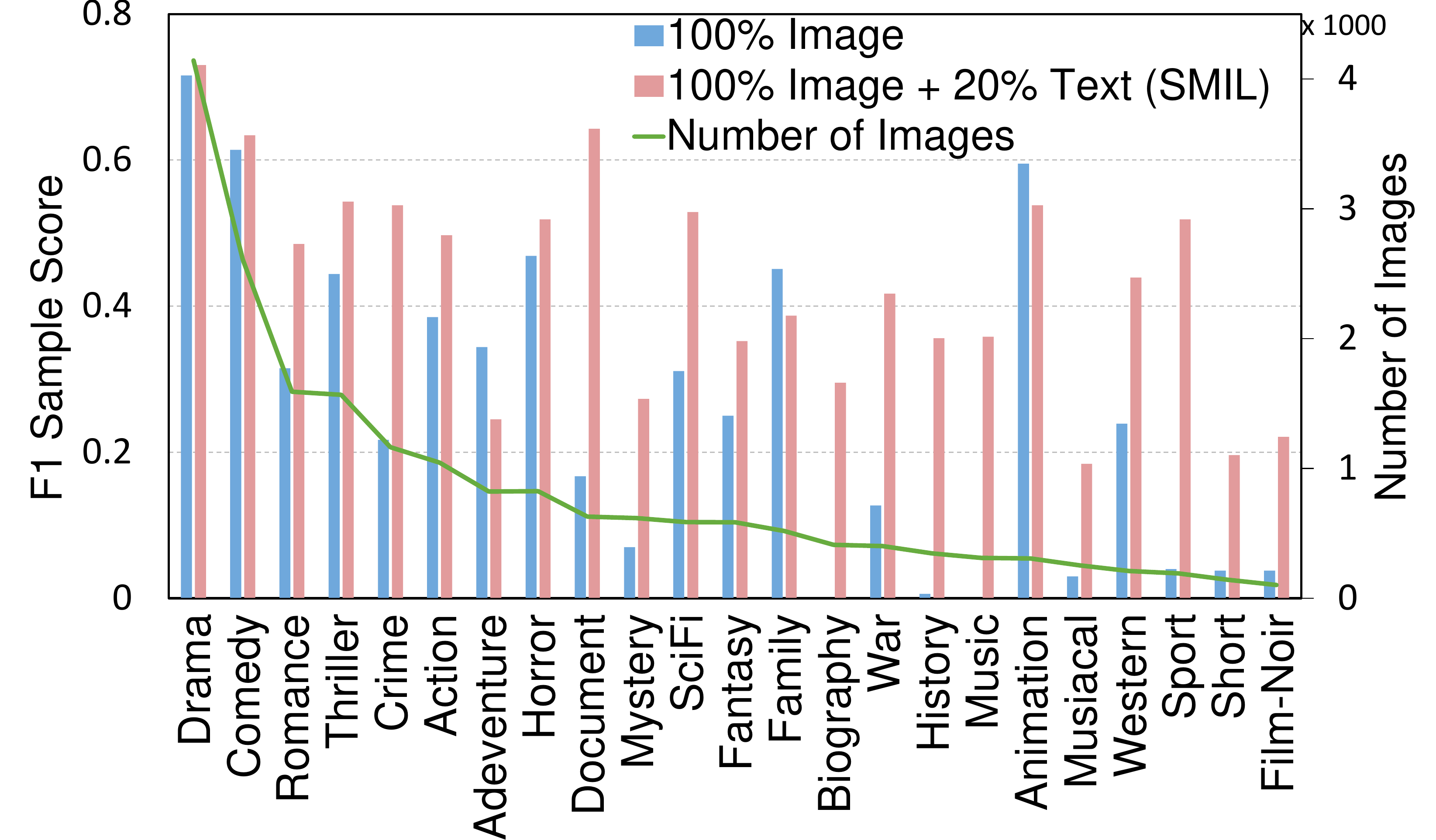}
    \caption{F1 Samples score of each movie genre on the \textit{MM-IMDb} dataset for the lower-bound baseline (\textit{blue}) and SMIL (\textit{red}). The number of image samples for each movie genre is indicated by the {green} line.}
    \label{fig:MM-IMDb}
\end{figure}

 We further show the effect of multimodal learning for different classes of MM-IMDb when $\eta = 20\%$ in Figure~\ref{fig:MM-IMDb}. First, our method (shown as red bars) can largely improve the model performance even on the tailed genres, such as \textit{Sport} and \textit{Film-Noir}, while the model trained only using images (shown as blue bars) can hardly predict the classes with less training samples. Second, an interesting phenomenon in Figure~\ref{fig:MM-IMDb} is that text modality will slightly decrease the performance of movie genres like \textit{Family} and \textit{Animation}. The possible reason is that there is a large overlap between genres of family and animations. As a result, text modality may enforce the model to learn the shared knowledge between these two genres, which reduces the discrepancy and decrease the accuracy.



{\bf Visualization of embedding space.}
We visualize the embedding space of three genres in MM-IMDb in Figure~\ref{fig:tsne}, and observed that our approach can effectively disentangle the latent embedding of the three genres, while the model learned only from image modality cannot. Besides, Our method is efficient when modality is severely missing. Form Figure 5, we see that our model trained using only $10$\% text modality is comparable to a model trained using $100$\% text modality.

{\bf Justification of symbol `-' used in Table 1, 2.}
We use the ‘-’ symbol for two reasons. First, not applicable. Lower-Bound only requires image modality for training, so it is not applicable to report a Lower-Bound result trained using both image and text. Second, not necessary. For example, in table 1, MVAE trained without missing modality (100\% image + 100\% text) achieves acc = 58.5\%. In comparison, our model trained with severely missing modality (100\% image + 10\% text) achieves acc = 60.7\%. So it is not necessary to train MVAE under severely missing modality.

\begin{figure}[t]
    \centering
    \includegraphics[width=\linewidth]{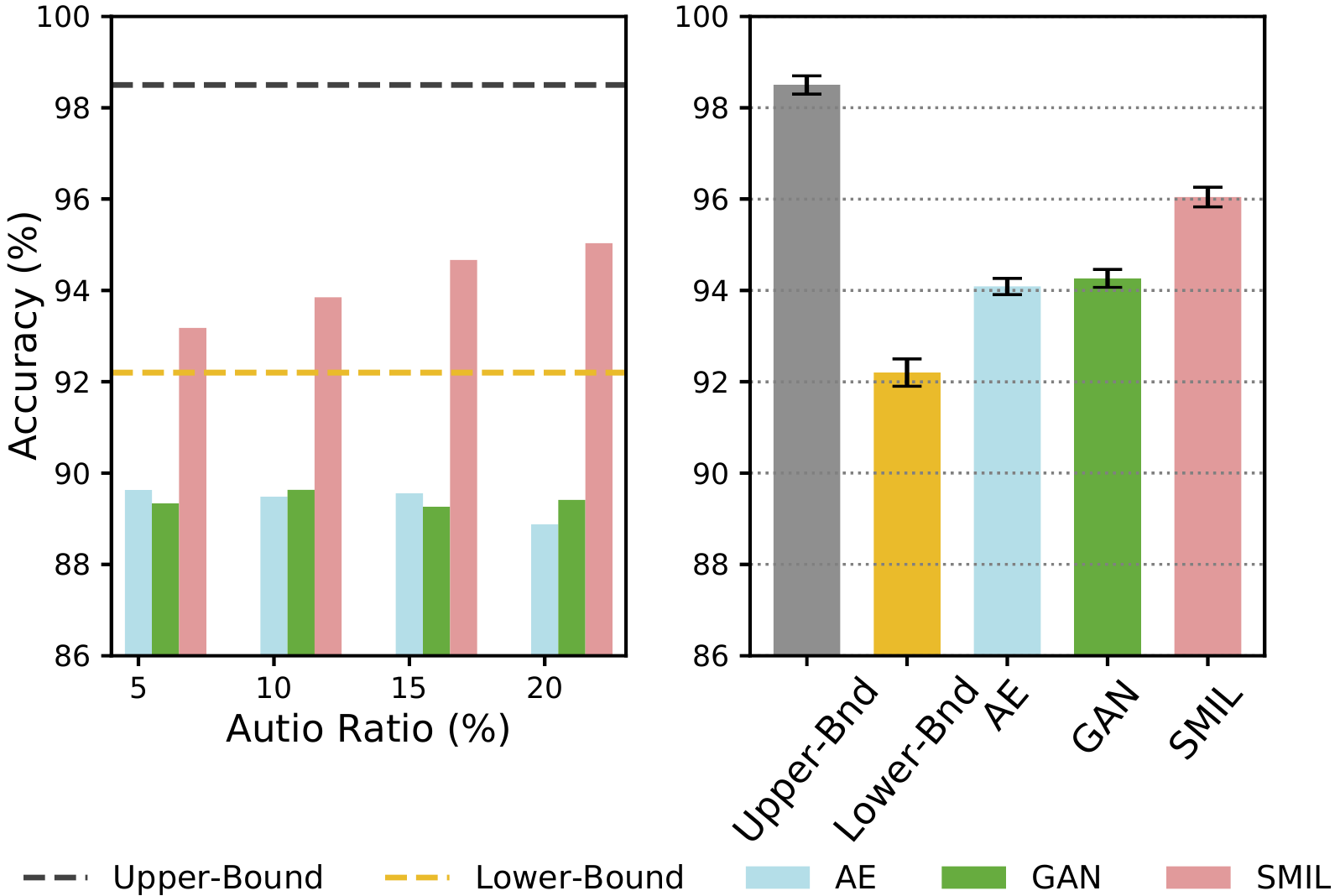}
    \caption{Classification accuracy (\%) on \textit{avMNIST} with two missing patterns. \textit{Left}: training with 100\% Image + $\eta$\% Audio and testing with Image Only. \textit{Right}: training with 100\% Image + $\eta$\% Audio and testing with Image + Audio.}
    \label{fig:avmnist}
\end{figure}

\subsection{Flexibility with Different Missing Patterns}
\textbf{Conclusion:} \textit{ Our method shows flexibility in handling various missing patterns: (1) full or missing modality at training; and (2) full or missing modality at test time.}

{\bf Implementation details.}  Our network contains two modality-specific feature extractors and a few FC layers. We use LeNet-5 to extract features for image modality, and a modified LeNet-5 to extract audio features. Extracted features are then fused through concatenation and sent into FC layers to perform classification. For the training process, we use Adam optimizer with a batch size of $64$ and train the networks for $15,000$ iterations with a learning rate of $10^{-3}$ for both inner- and outer- loop of meta-learning.

{\bf Setting of missing pattern.} For the avMNIST dataset, the missing modality problem only happens to audio modality. We are interested in two different missing patterns: (1) training with 100\% Image + $\eta$\% Audio and testing with Image Only; (2) training with 100\% Image + 20\% Audio and testing with Image + Audio. In this section, we show that our approach can flexibly handle these two missing patterns.

{\bf Missing pattern 1: testing with image only.}
Figure~\ref{fig:avmnist} (left) shows the classification accuracy under different audio ratio. We see that our approach can successfully handle testing with image modality only, but baseline methods such as AE and GAN fail in this scenario. As can be seen, when $\eta=20\%$, SMIL is 6.7\% higher than the generative-based method, and 3.3\% higher the Lower-Bound.  We argue that the failure of baseline methods is mainly due to the bias of the reconstructed missing modality. In single modality testing, the method is required to generate the missing modality conditioned on the available modality. The baseline method does not consider the bias of the reconstructed missing modality. In contrast, our method can leverage learned meta-knowledge to generate an unbiased missing modality. Besides, in situations where audio modality is missing severely (\textit{i.e.}, $\eta=5\%$), The classification accuracy of our method is $1.10$\% higher than the lower bound. The improvement demonstrates clear advantages of our model under severely missing modality.

{\bf Missing pattern 2: testing with image and audio.}
Figure~\ref{fig:avmnist} (right) shows the result of our approach dealing with full modality testing. We observe that our method still performs the best. It outperforms the Lower-Bound by 4.3\% and the generative-based method (AE and GAN) by 2.1\%. Moreover, under different missing patterns, SMIL is consistently better than AE and GAN. When switching testing patterns from two modalities to a single modality, AE and GAN have a 5.6\% performance drop, while SMIL only has a 1.0\% performance drop.  


\subsection{Ablation Study}

\begin{table}[t]
\centering
\resizebox{\linewidth}{!}{
    \begin{tabular}{lcccc}
    
    \toprule
    \multirow{2}{*}{Method} & \multicolumn{2}{c}{F1 Samples $\uparrow$} & \multicolumn{2}{c}{F1 Micro $\uparrow$}  \\ 
    \cmidrule{2-3}
    \cmidrule{4-5}
        & 10\% & 20\% & 10\% & 20\%      \\
    \midrule
     SMIL w/o K-means     & {0.482} &0.535  & {0.485}  & {0.530} \\
     \midrule
   SMIL w/o Regularization    & {0.469} & 0.521 & {0.472} & {0.530} \\ 
   \midrule
    SMIL w/ Fixed Gaussian     & {0.475} & 0.495 & {0.479} & {0.502}      \\
    SMIL w/ Deterministic      & {0.474} &0.527  & {0.477}  & {0.533} \\ 
    \midrule
    SMIL (Full)    & \textbf{0.492} & \textbf{0.541} &\textbf{0.495} &\textbf{0.546}\\
    \bottomrule
    \end{tabular}}
    \caption{Ablation study on the effect of modality reconstruction, feature regularization, and Bayesian inference on \textit{MM-IMDb} under two text modality ratios (10\% and 20\%). }\label{tab:ablation}
\end{table}
We conduct the ablation analysis on the MM-IMDb dataset to evaluate the effectiveness of the missing modality reconstruction, feature regularization, and Bayesian Inference. We show the results in Table~\ref{tab:ablation}.

{\bf Effectiveness of missing modality reconstruction.} In Section~\ref{subsec:missingModality}, we use reconstruction network to generate weights for missing modality reconstruction. Here we denote the method that uses the reconstruction network to directly generate the feature of missing modality as \textit{SMIL w/o K-means}, which has worse performance and proves the necessity of K-Means for reconstruction.

{\bf Effectiveness of feature regularization.} In Section~\ref{subsec:regularization}, we introduce feature regularization. Here we denote the method without feature regularization as \textit{SMIL w/o Regularization}. The performance of \textit{SMIL w/o Regularization} is inferior to \textit{SMIL (Full)}, which verifies conducting multimodal learning on $D$ without regularization  leads to a sub-optimal model. The superior performance of the regularized model is essential to the explicit objective of reducing discrepancy.
 
{\bf Effectiveness of Bayesian inference.} In Section~\ref{subsec:bayesianInference}, we introduce the Bayesian Meta-Learning Framework. In this section, we compare it with two variants. \textit{SMIL w/ Fixed Gaussian}: We fix the  distribution of feature regularization to a Gaussian distribution, which is  $\mathcal{N}(\textbf{0},\textbf{I})$; \textit{SMIL w/ Deterministic}:  The missing modality construction and feature regularization is deterministic so the sampling in Eqn.~\ref{eqn:6} is removed. These two variants are inferior to Bayesian inference.The results show the superiority of Bayesian Meta-Learning framework.

\begin{figure}[t]
    \centering
    \includegraphics[width=\linewidth]{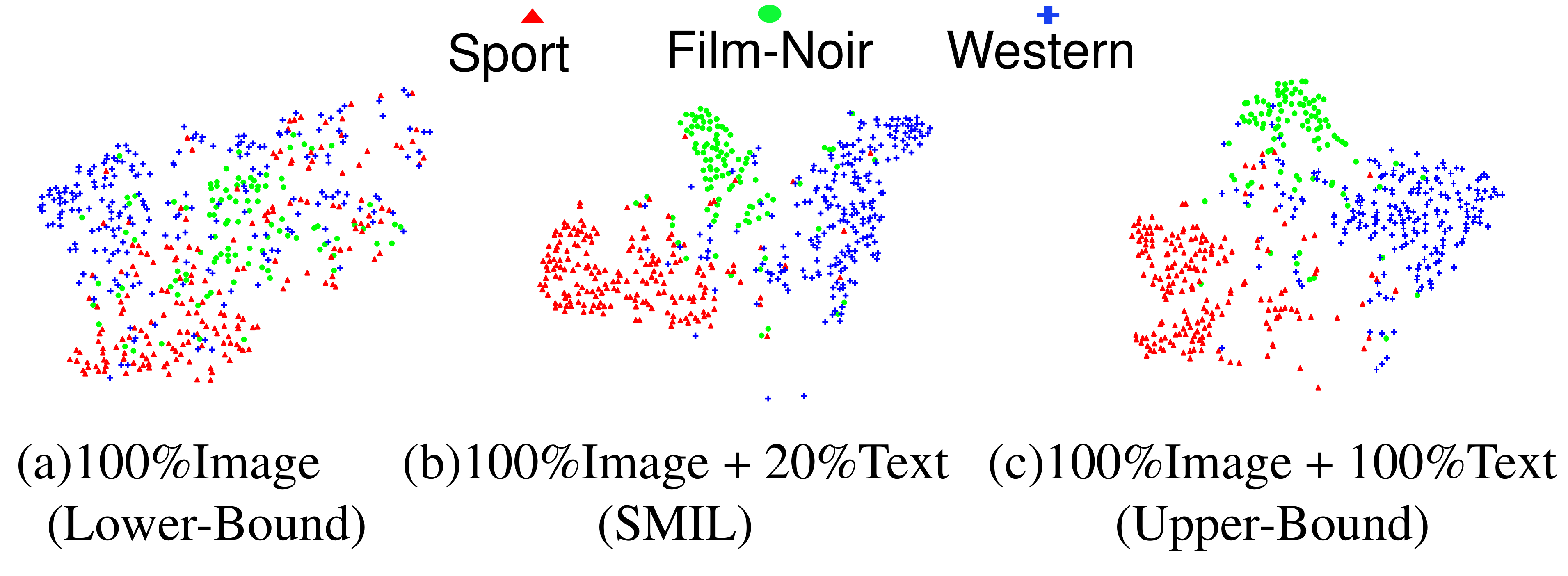} 
    \caption{t-SNE visualization for embeddings of the lower-bound baseline~(a), SMIL~(b), and upper-bound baseline~(c) on the \textit{MM-IMDb} dataset. Three movie genres, including \texttt{Sport}, \texttt{Film-Noir}, and \texttt{Western} are visualized.}
    \label{fig:tsne}
\end{figure}

\section{Conclusion}
In this paper, we address a challenging and novel problem in multimodal learning: multimodal learning with severely missing modality. We further propose a novel learning strategy based on the meta-learning framework. This framework tackles two important perspectives: missing modality reconstruction (flexibility) and feature regularization (efficiency). We apply the Bayesian meta-learning framework to infer the posterior of them and propose a variational inference framework to estimate the posterior.

In the experiments, we show that our model outperforms the generative method significantly on three multimodal datasets. Further analysis on the results shows that involving modality reconstruction and feature regularization can effectively handle the missing modality problem and flexible to various missing patterns. We believe that our work makes a meaningful step towards the real-world application of multimodal learning where partial modalities are missing or hard to collect.

\section{Acknowledgements}
This work is partially supported by the Data Science Institute (DSI) at University of Delaware and Snap Research.

\small
\bibliography{references.bib}

\end{document}